\documentclass[10pt,twocolumn,letterpaper]{article}

\usepackage{3dv}
\usepackage{times}
\usepackage{epsfig}
\usepackage{graphicx}
\usepackage{amsmath}
\usepackage{amssymb}
\usepackage{soul}
\usepackage{mathtools}
\usepackage{booktabs}
\usepackage[font=small,skip=4pt]{caption}
\usepackage[font=small]{subfig}
\usepackage{xcolor, colortbl}
\usepackage{multirow}
\usepackage{tabularx}
\usepackage[super]{nth}
\newcolumntype{Y}{>{\centering\arraybackslash}X}
\usepackage{enumitem}
\usepackage{algorithm}
\usepackage{algorithmic}

\usepackage{color,soul}

\soulregister\cite7
\soulregister\ref7
\soulregister\pageref7

\definecolor{mygreen}{RGB}{0, 176, 80}
\definecolor{myerrorblue}{RGB}{49,  54, 149}
\definecolor{myerrorred}{RGB}{215,  48,  39}

\colorlet{lightgrey}{gray!25}
\colorlet{lightred}{red!25}
\colorlet{lightgreen}{green!25}
\colorlet{lightyellow}{yellow!25}

\threedvfinalcopy 

\def\threedvPaperID{167} 

\ifthreedvfinal\fi
\begin{document}

\title{Nighttime Stereo Depth Estimation using Joint Translation-Stereo Learning: Light Effects and Uninformative Regions}

\author{Aashish Sharma$^{1}$, Loong-Fah Cheong$^{1}$, Lionel Heng$^{3}$, and Robby T. Tan$^{1,2}$\\
$^1$National University of Singapore, $^2$Yale-NUS College, $^3$DSO National Laboratories\\
{\tt\small 
aashish.sharma@u.nus.edu, eleclf@nus.edu.sg, lionel.heng@ieee.org, robby.tan@nus.edu.sg}}

\maketitle

\def\thefootnote{$\dagger$}\footnotetext{This work is supported by MOE2019-T2-1-130.}\def\thefootnote{\arabic{footnote}}
\maketitle

\begin{abstract}
	Nighttime stereo depth estimation is still challenging, as assumptions associated with daytime lighting conditions do not hold any longer.
	Nighttime is not only about low-light and dense noise, but also about glow/glare, flares, non-uniform distribution of light, etc. 
	One of the possible solutions is to train a network on night stereo images in a fully supervised manner. However, to obtain proper disparity ground-truths that are dense, independent from glare/glow, and have sufficiently far depth ranges is extremely intractable. 
	To address the problem, we introduce a network joining day/night translation and stereo. In training the network, our method does not require ground-truth disparities of the night images, or paired day/night images. 
	We utilize a translation network that can render realistic night stereo images from day stereo images. 
	We then train a stereo network on the rendered night stereo images using the available disparity supervision from the corresponding day stereo images, and simultaneously also train the day/night translation network. 
	We handle the fake depth problem, which occurs due to the unsupervised/unpaired translation, for light effects (e.g., glow/glare) and uninformative regions (e.g., low-light  and saturated regions), by adding structure-preservation and weighted-smoothness constraints.
	Our experiments show that our method outperforms the baseline methods on night images.
\end{abstract}

\begin{figure}[t!]
	\vspace{-0.1in}
	\captionsetup[subfloat]{labelformat=empty}
	\captionsetup[subfloat]{farskip=2pt}
	\centering
	\subfloat[Input Night Image]{\includegraphics[width=0.235\textwidth]{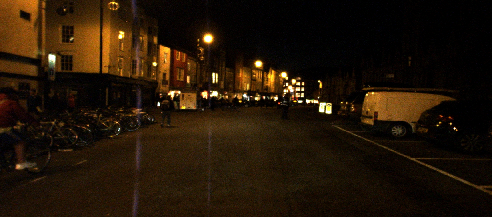}}\hfill
	\subfloat[Boosted Night Image]{\includegraphics[width=0.235\textwidth]{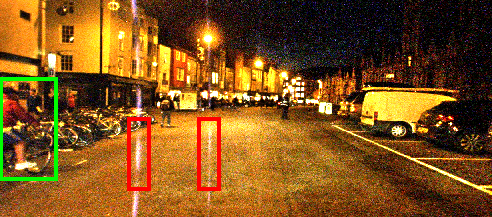}}\hfill\\
	\subfloat[Sparse Depth Map]{\includegraphics[width=0.235\textwidth]{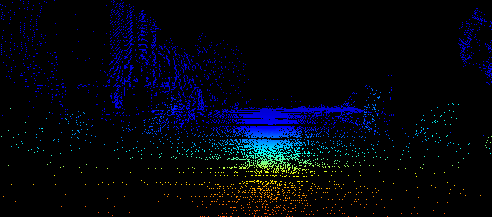}}\hfill
	\subfloat[ PSMNet~\cite{chang2018pyramid} ]{\includegraphics[width=0.235\textwidth]{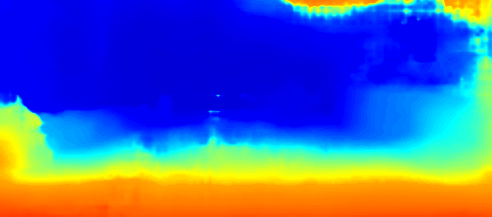}}\hfill\\
	\subfloat[ Joint-SS~\cite{sharma2018into}]{\includegraphics[width=0.235\textwidth]{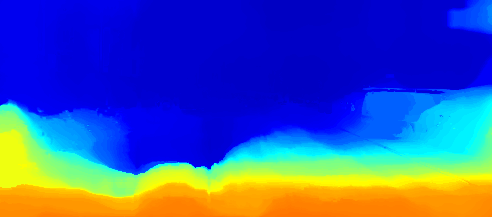}}\hfill
	\subfloat[\textbf{Our Result}]{\includegraphics[width=0.235\textwidth]{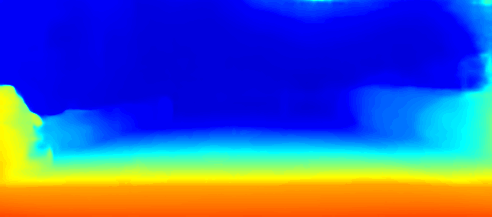}}\hfill\\
	
	\caption{For the input night image from the Oxford dataset~\cite{maddern20171}, compared with the existing baselines, PSMNet~\cite{chang2018pyramid} and Joint-SS~\cite{sharma2018into}, our result is more accurate and robust to problems such as flares (see the areas marked in red in the boosted night image). It is sharper and has better depth discontinuities (observe the cyclist marked in green in the boosted night image). Note that, the boosted night image (obtained with~\cite{guo2017lime}) and the sparse depth map are shown for the sake of visualization.}\label{figure_into}
	\vspace{-0.125in}
\end{figure}

\section{Introduction}
Depth from stereo is an important research area in computer vision and  essential for many real-world applications. 
%
%
Progress has been made significantly in the field~\cite{laga2020survey}, yet despite this,  depth from stereo under nighttime conditions is still an unexplored area. Conventional non-learning based methods (e.g., \cite{hirschmuller2005accurate,zhang2015meshstereo}) rely on brightness and gradient constancy assumptions, which are severely violated in the nighttime conditions due to low-light, noise, flares, glow/glare, varying illumination, etc. \cite{sharma2018into}. 

Deep learning based methods trained on a daytime dataset (e.g. \cite{chang2018pyramid,kendall2017end}) are unable to work properly on night images, as shown in Fig.~\ref{figure_into}. This is because night images contain flares, dense noise, glow/glare, etc., which are not present in the daytime training data. One possible solution is to train the network on night images in a fully supervised manner. However, to obtain the disparity ground-truths of night images is not trivial. Maddern et al. \cite{maddern20171} create a dataset of 3D point clouds using LIDAR sensors and the corresponding images captured using normal cameras. From this dataset, we can obtain the depths of the respective images. However, these depths are sparse, and cannot capture independently moving objects (e.g., moving cars, bikes, pedestrians, etc.) as well as distant objects (for e.g., see the sparse depth map in Fig.~\ref{figure_into}). In other words, if we use these depths as ground-truths for training, they will be considerably noisy.

In this paper, our goal is to address the problem of depth from stereo for night images. Our idea is to employ a network that performs image translation from day-to-night (and night-to-day), and stereo matching simultaneously, with the two processes working to benefit each other. 
We train our translation network to transform an original night image pair to a rendered day image pair, and also to transform an original day image pair to a rendered night image pair. Simultaneously, we train our stereo network to compute the disparity map of the rendered night image pair, guided by the disparity map computed from the original day image pair. Our method uses no direct supervision and thus does not require the depth ground-truths of the input night images, or paired day/night images.

Unlike existing image translation networks (e.g., \cite{zhu2017unpaired,liu2017unsupervised,hoffman2017cycada}),
our translation network utilizes a stereo-consistency constraint and translates the stereo pair together, which reduces the inconsistencies in the translated pair and helps to generate a more robust depth result. Moreover, our method is designed to deal with the fake depth problem. Due to unsupervised/unpaired translation, objects in one domain for which the network cannot find the corresponding objects in the other domain, can be transformed into fake image structures.
Night mages contain light effects such as glow/glare that are not present in day images, hence, the network will not be able to find their counterpart objects in day images, rendering fake structures. This also happens for uninformative regions such as regions under dark noise (or low-light regions) and saturated regions. The reason is simply because there is no information in these regions for the network to transform them properly. 

The presence of fake image structures can provide inaccurate/fake depths for stereo, and hence we intend to also address this problem. 
First, in the training stage, for handling regions containing glow/glare light effects, we use a structure-preservation constraint. This is based on the idea that beyond the light effects, there is structural similarity between the input night (or day) images and the rendered day (or night) images.  
Second, in the testing stage, for handling uninformative regions, we finetune our method on the testing night pair (i.e., test-time optimisation) using a self-supervised weighted-smoothness constraint. The basic idea is that in the uninformative regions, we propagate depths from the neighboring informative regions that have more correct depths. As shown in Fig.~\ref{figure_into}, our method outperforms the state-of-the-art methods in both general stereo vision \cite{chang2018pyramid} and nighttime stereo vision \cite{sharma2018into}.

In summary, our contributions are as follows:
\begin{itemize}[leftmargin=*]
	\setlength\itemsep{-0.03in}
	\vspace{-0.1in}
	\item We introduce an end-to-end learning-based joint translation and stereo network to address the problem of nighttime stereo depth estimation. Our method uses no direct supervision and does not need ground-truth disparities for training on night images. 
	\item We introduce a stereo-consistency constraint into the translation network to ensure that it generates a rendered pair stereo-consistently, namely, the stereo images in a rendered pair contain consistent image structures.
	\item  To our knowledge, our method is the first to address the fakeness problem that occurs in unsupervised/unpaired image translation. For night images, this problem arises for glow/glare light effects and uninformative regions. To address the problem, we propose to add a structure-preservation constraint in the training stage for handling glow/glare regions; and a self-supervised weighted-smoothness constraint in the testing stage for handling the uninformative regions. 
	\vspace{-0.02in}
\end{itemize}
Our experiments show that our method provide sharper and more accurate depth results, particularly for robustness to glow/glare, streak-like flares, low-light regions, etc. 

\begin{figure*}[t!]
	\vspace{-0.1in}
	\captionsetup[subfloat]{labelformat=empty}
	\captionsetup[subfloat]{farskip=2pt}
	\centering
	\subfloat{\includegraphics[width=1.0\textwidth,height=8cm]{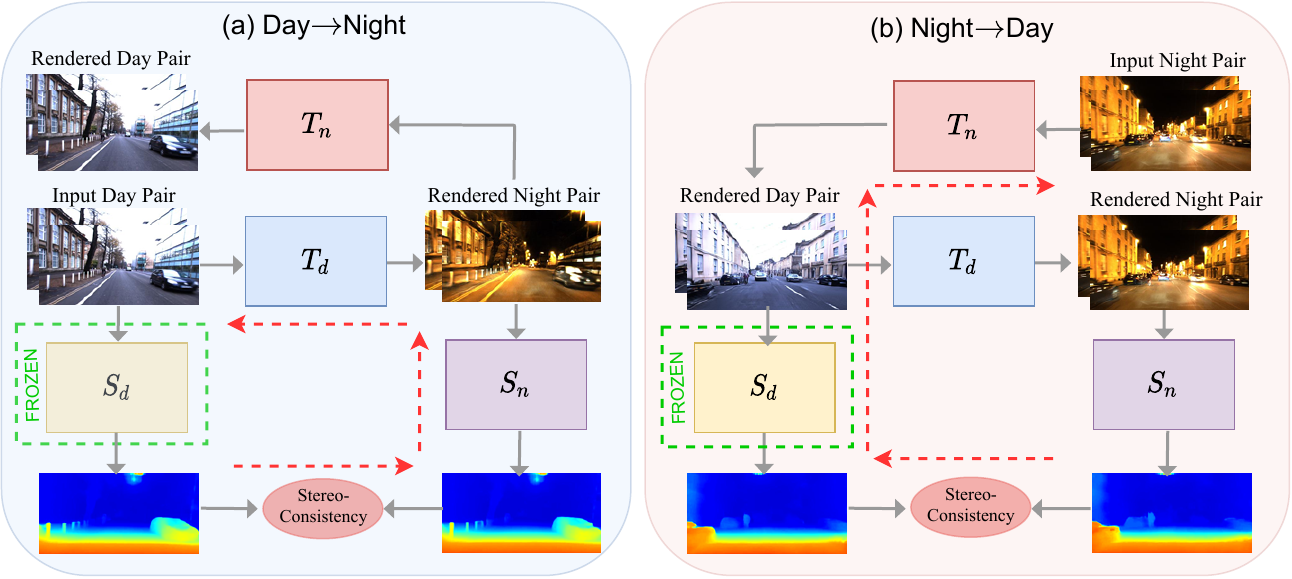}}
	\caption{The two concurrent cycles of our network are shown on the left and right side respectively. For illustrating the stereo-consistency constraint, discriminators and the other losses are omitted. In the day$\rightarrow$night cycle (see (a)), enforcing the stereo-consistency constraint, we jointly train $T_d$ and $S_n$. This ensures that $T_d$ generates stereo-consistent translations, while $S_n$ is getting trained on them. In the night$\rightarrow$day cycle (see (b)), we train $T_n$ with the stereo-consistency constraint. To elucidate the usage of the stereo-consistency constraint, we have red-dotted arrows showing the associated back-propagation directions in the two cycles. }\label{figure_our_model}
	\vspace{-0.1in}
\end{figure*}

\section{Related Work}
Depth from  stereo has been investigated extensively, and for a comprehensive review we refer to Laga et al.'s \cite{laga2020survey}. 
Nighttime stereo however, is a relatively less investigated area given its numerous  challenges. Hence,  despite its significance, there are only few methods   that have attempted to address the problem~\cite{heo2007simultaneous,jiao2017joint,sharma2018into}. Both Heo et al.~\cite{heo2007simultaneous} and Jiao et al.~\cite{jiao2017joint} assume the noise distribution to be Gaussian, and thus do not generalize well on real night images. Moreover, Heo et al.'s~\cite{heo2007simultaneous} is  limited to only low Gaussian noise levels $(\text{st.d.}\le25)$. Recently, an optical flow estimation method under low-light conditions is proposed~\cite{flowdark}. However, this method uses RAW images as input (not RGB images), that limits its application since RAW images might not be available in many practical scenarios.

Sharma and Cheong's \cite{sharma2018into} attempts to address the nighttime stereo vision problem by optimizing a joint structure-stereo model that performs structure extraction and stereo estimation jointly. There is no explicit assumption on the noise distribution, and hence, the method generalizes well on night images, and it is the current state-of-the-art method. However, owing to its high optimization complexity, it is computationally inefficient, and still suffers from other nighttime problems such as flares, glow/glare, etc. 

Our method has a similar translation component to that used in Hoffman et al.'s \cite{hoffman2017cycada} but with a few key differences: 1) Our translation network includes the stereo-consistency constraint and translates stereo pairs together, while \cite{hoffman2017cycada} models semantic-consistency on individual images. 2) Our translation network includes structure-preservation and weighted-smoothness constraints to address the fake depth problem for glow/glare light effects and uninformative regions in night images. While~\cite{hoffman2017cycada} has no explicit component to deal with the fake depth (or fake semantics) problem, which as mentioned, can occur due to unsupervised/unpaired learning of the translation process.  

\section{Our Method}
\label{section_our_method}
We have two input domains - day and night stereo images. We represent the stereo image pairs by $(X^l_d, X^r_d)$ and $(X^l_n, X^r_n)$, where the subscript defines the domains ($d$ = day, and $n$ = night) and the superscript defines the left/right image view. A stereo pair drawn from a set is represented by small letters: $(x^l_d, x^r_d)$  for a day pair and $(x^l_n, x^r_n)$ for a night pair. The two domains are uncorrelated/unpaired, i.e., for any day stereo pair, we do not have its exact corresponding night stereo pair (and vice-versa). Moreover, we assume that we have the disparity ground-truths available for the day stereo images, represented by $Y^l_d$. However, we do not have the ground-truths for the night stereo images. Our goal is to estimate the disparity map of a given night image pair, $(x^l_n, x^r_n)$.

To achieve the goal, we employ two networks: translation and stereo networks, as shown in Fig.~\ref{figure_our_model}. Our translation network is represented by $\{(T_d, D_d), (T_n, D_n)\}$  with $T$ and $D$ being the generator and discriminator networks respectively. Our stereo part contains two identical networks, $\{S_d, S_n\}$, each based on the stacked hourglass model~\cite{chang2018pyramid}. We use $T_d$ to perform the day$\rightarrow$night translation, and $T_n$ to perform the night$\rightarrow$day translation. The two translations occur in two concurrent cycles as shown in Fig.~\ref{figure_our_model}.

In our method, we assume that $S_d$ is already pre-trained on the day stereo images and $S_n$ is initialized with $S_d$'s trained weights. Since we do not need $S_d$ to learn any further for the day domain, we keep it completely frozen during the training process. For $S_n$, in the day$\rightarrow$night cycle, we train it together with $T_d$ on the translated night pairs using the disparity supervision from their input day counterparts (see Fig.~\ref{figure_our_model}a). However, we do not train $S_n$ in the night$\rightarrow$day cycle, since we have no disparity ground-truths available for the input night pairs and only train $T_n$ (see Fig.~\ref{figure_our_model}b). The details of the steps are discussed next.

\begin{figure*}[t!]
	\vspace{-0.1in}
	\centering
	\subfloat{\includegraphics[width=1.0\textwidth]{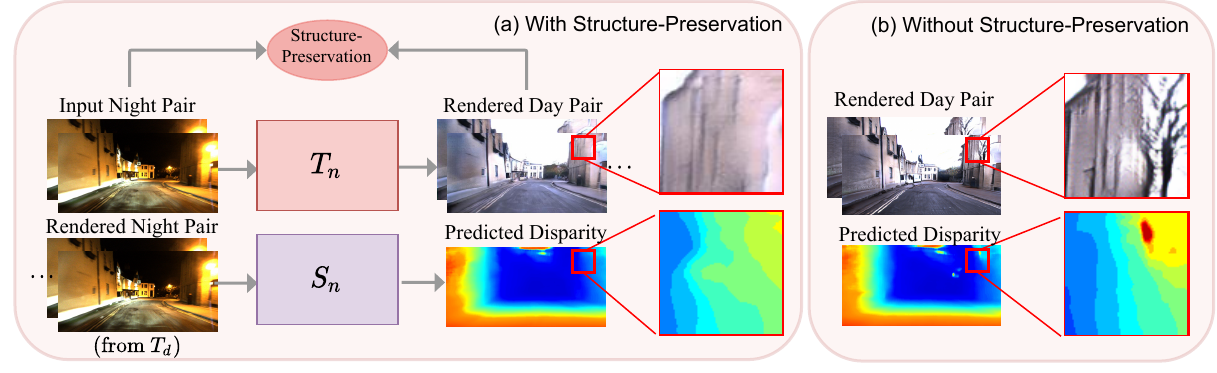}}
	\caption{Handling fake depths for regions containing light effects such as glow/glare by using structure-preservation constraint. Note that, the disparity patches are scaled for visualization. Zoom-in for better display.}\label{figure_our_model2}
	\vspace{-0.1in}
\end{figure*}

\subsection{Learning Nighttime Depth Estimation}
\label{subsec_learning_nighttime_depth} 
We pre-train $S_d$ on the day pairs, $(X^l_d, X^r_d)$, using their corresponding disparity ground-truths, $Y^l_d$. For this, given $(x^l_d, x^r_d)$ drawn from $(X^l_d, X^r_d)$ and $y^l_d$ drawn from  $Y^l_d$, the loss is expressed as:
\vspace{-0.05in}
\begin{equation}
\mathcal{L}_\text{disp}(S_d) = \mathbb{E}_{(x^l_d, x^r_d), y^l_d}\Bigg[\sum_{k=1}^{3}w_k\cdot \lVert {y_k}^l_d - y^l_d\rVert_1\Bigg]\label{eqn_stereo_S_d},
\end{equation}
where $S_d[(x^l_d, x^r_d)] = \{{y_k}^l_d\}_{k=1,2,3}$ represent the three prediction branches from the stacked hourglass model of $S_d$. $w_k$ is the weighting factor of the corresponding  term. Having pre-trained $S_d$, we freeze its weights completely. The stereo network $S_n$ is initialized with the pre-trained weights of $S_d$, and the translation network, $\{(T_d, D_d), (T_n, D_n)\}$, are initialized randomly.

\subsubsection{Discriminative Loss}
\vspace{-0.05in}
In the day$\rightarrow$night cycle, given any day pair  $(x^l_d, x^r_d)$ and any night pair  $(x^l_n, x^r_n)$, the rendered night pair is generated by the daytime generator: $(z^l_n, z^r_n) = T_d[(x^l_d, x^r_d)]$, where $(T_d, D_d)$ are trained using the following loss functions:
\vspace{-0.01in}
\begin{align}
\mathcal{L}_\text{GAN}(T_d) = &
\mathbb{E}_{(x^l_d, x^r_d)(x^l_n, x^r_n)}\big[(D_d[(z^l_n, z^r_n)] - \phantom{zzzzz}\nonumber \\
& \phantom{zzzzzzzzzzzzzz} D_d[(x^l_n, x^r_n)] - 1)^2\big],\label{eqn_GAN_T_d} \\
\mathcal{L}_\text{GAN}(D_d) = &
\mathbb{E}_{(x^l_d, x^r_d)(x^l_n, x^r_n)}\big[(D_d[(x^l_n, x^r_n)] 
- \phantom{zzzzz}\nonumber \\
& \phantom{zzzzzzzzzzzzzz} D_d[(z^l_n, z^r_n)] - 1)^2\big],\label{eqn_GAN_D_d} 
\end{align}
where we use the least-squares discriminative loss \cite{mao2017least} adapted to the relativistic formulation~\cite{jolicoeur2018relativistic} to stabilize the training process. For the night$\rightarrow$day cycle, we obtain the loss functions $\mathcal{L}_\text{GAN}(T_n)$ and $\mathcal{L}_\text{GAN}(D_n)$ from Eqs.~(\ref{eqn_GAN_T_d}) and (\ref{eqn_GAN_D_d}), by swapping the domain labels from day to night. 

\subsubsection{Cycle-Consistency Loss}
\vspace{-0.05in}
In the day$\rightarrow$night cycle, we can further translate the rendered night pair $(z^l_n, z^r_n)$ to the rendered day pair $(\widetilde{x}^l_d, \widetilde{x}^r_d)$ using the nighttime generator:
$(\widetilde{x}^l_d, \widetilde{x}^r_d)= T_n[(z^l_n, z^r_n)]$.
Based on the cycle-consistency constraint \cite{zhu2017unpaired}, we expect that the rendered day pair $(\widetilde{x}^l_d, \widetilde{x}^r_d)$ should be the same as the original day pair $(x^l_d, x^r_d)$. Thus we can define a cycle-consistency loss function to train the daytime generator, $T_d$:
\vspace{-0.01in}
\begin{equation}
\mathcal{L}_\text{cyc-con}(T_d) = \mathbb{E}_{(x^l_d, x^r_d)}[\lVert \widetilde{x}^l_d - x^l_d\rVert_1 + \lVert \widetilde{x}^r_d - x^r_d\rVert_1].\label{eqn_cyc_con_T_d}
\end{equation}
For the night$\rightarrow$day cycle, we can similarly obtain $\mathcal{L}_\text{cyc-con}(T_n)$ from Eq.~(\ref{eqn_cyc_con_T_d})  by swapping the domain labels from day to night.

\subsubsection{Stereo-Consistency Loss}
\vspace{-0.05in} 	
As for the stereo-consistency constraint, the idea is that the disparity computed from  the rendered night pair $(z^l_n, z^r_n)$
should be the same as the disparity computed from
the original day pair $(x^l_d, x^r_d)$, since they are originated from the same images. 
As a result, in the day$\rightarrow$night cycle, we can train $T_d$ and $S_n$ based on this stereo-consistency constraint. 
Specifically, using $S_d[(x^l_d, x^r_d)] = \{{y_k}^l_d\}_{k=1,2,3}$ as disparity supervision, and $S_n[(z^l_n, z^r_n)] = \{\widehat{y_k}^l_n\}_{k=1,2,3}$ as disparity prediction, we have the following stereo-consistency loss function to train  $(T_d, S_n)$:
\vspace{-0.1in}
\begin{equation}
\mathcal{L}_\text{ste-con}(T_d, S_n) = \mathbb{E}_{(z^l_n, z^r_n), {y_k}^l_d}\!\Bigg[\sum_{k=1}^{3}\lVert \widehat{y_k}^l_n - {y_k}^l_d)\rVert_1\!\Bigg].\label{eqn_ste_con_T_d_S_n}
\end{equation}

The stereo-consistency loss in the night$\rightarrow$day cycle is similar to Eq.~(\ref{eqn_ste_con_T_d_S_n}), where we expect the disparity computed from the rendered day pair, $(z^l_d, z^r_d)$, to be the same as the disparity computed from the original night pair, $(x^l_n, x^r_n)$. However, there are two important changes. First, we do not train $S_n$ this time as there is no disparity ground-truth available for the input night image pair $(x^l_n, x^r_n)$. 
Second, for the stereo-consistency loss, we compute the disparity from the rendered night pair $(\widetilde{x}^l_n, \widetilde{x}^r_n)$, instead of the original night pair $(x^l_n, x^r_n)$.

The reason is because in the beginning of the training process, the rendered night  pair in the day$\rightarrow$night cycle, $(z^l_n, z^r_n)$, is different from the original night pair, $({x}^l_n, {x}^r_n)$, yet it is more similar to the rendered night pair in the night$\rightarrow$day cycle, $(\widetilde{x}^l_n, \widetilde{x}^r_n)$. Since, both of them are synthetically rendered pairs. 
Using the rendered night pairs, we can get a more stable learning process (as shown in our ablation study). Therefore, using $S_n[(\widetilde{x}^l_n, \widetilde{x}^r_n)] = \{\widetilde{y_k}^l_n\}_{k=1,2,3}$ as disparity supervision, and $S_d[(z^l_d, z^r_d)] = \{\widehat{y_k}^l_d\}_{k=1,2,3}$ as disparity prediction, we have the following stereo-consistency loss function to train  $T_n$:
\vspace{-0.05in}
\begin{equation}
\mathcal{L}_\text{ste-con}(T_n) = \mathbb{E}_{(z^l_d, z^r_d), \widetilde{y_k}^l_n}\Bigg[\sum_{k=1}^{3}\lVert \widehat{y_k}^l_d - \widetilde{y_k}^l_n)\rVert_1\Bigg].\label{eqn_ste_con_T_n}
\end{equation}
As we can observe, the stereo-consistency constraint acts as the coupling link between the two processes (translation and stereo) in our joint network. 

\begin{figure*}[t!]
	\vspace{-0.1in}
	\centering
	\subfloat{\includegraphics[width=1.0\textwidth]{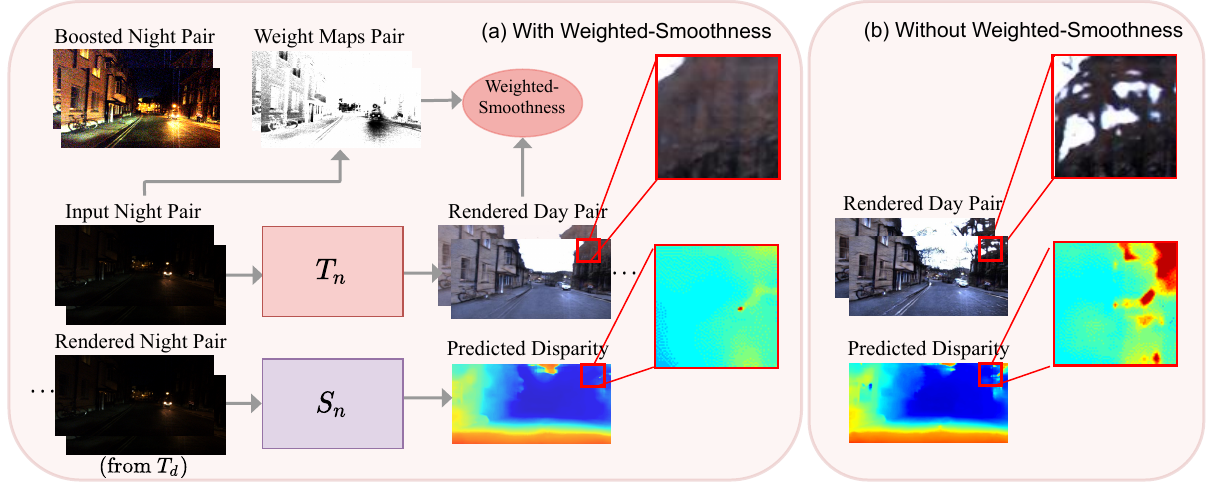}}
	\caption{Handling fake depths for uninformative regions such as low-light and saturated regions by using weighted-smoothness constraint. Note that, the disparity patches are scaled and the boosted night pair are shown for the sake of visualization. Zoom-in for better display.}\label{figure_our_model3}
	\vspace{-0.1in}
\end{figure*}

\subsection{Handling Fakeness in Nighttime Depths}
\label{subsec_handling_fake_depths} 

The discriminative loss is essential in image translation. However, due to unpaired translation, it can lead to the generation of fake image content/structures in the rendered images. 
In night images, this can happen for regions containing light effects such as glow/glare that have no counterparts in daytime; and for uninformative regions that do not have information in the first place, such as low-light regions (with zero or below dark noise pixel values) and saturated regions (with pixel values close to 255). 
For instance, in Fig.~\ref{figure_our_model2}b, we can observe that a glow/glare region is translated into fake trees; and in Fig.~\ref{figure_our_model3}b, where we can observe that some low-light regions are translated into fake/improper structures. 
This is a serious problem, since fake image structures can provide inaccurate/fake depths for a stereo system, which can be observed in the estimated depths in Figs.~\ref{figure_our_model2}b and~\ref{figure_our_model3}b. 

To address the fake depths problem, we propose two ideas. 
First, in the training stage, we handle the regions containing light effects such as glow/glare by using a structure-preservation constraint. 
Second, in the testing stage, we handle the uninformative regions by finetuning the network on the testing night pair $(x^l_n, x^r_n)$ using a self-supervised weighted-smoothness constraint.

\subsubsection{Structure-Preservation Loss}
Our structure-preservation constraint is based on the idea that structural features obtained from the input image pair should be similar to the structural features obtained from the corresponding rendered image pair~\cite{johnson2016perceptual}. 
To obtain the structural features for both night and day images, we employ a finetuned VGG-16 model, where we use a pre-trained VGG-16 (ImageNet~\cite{ImageNet}) and finetune it further on the night dataset, ExDark~\cite{ExDark}. 
ExDark provides night images with labeled objects, taken under varying illumination and light conditions. Thus the finetuned VGG-16 model can provide better structural features than the pre-trained VGG-16 model, and provides better performance (shown in our ablation study). 

Given an input night pair $(x^l_n, x^r_n)$ and the corresponding rendered day pair $(z^l_d, z^r_d)$, we define the structure-preservation loss in the night$\rightarrow$day cycle as:
\vspace{-0.01in}
\begin{equation}
\thickmuskip=0mu
\mathcal{L}_\text{str-pre}(T_n) = \mathbb{E}_{(x^l_n, x^r_n)}\Big[\big\lVert\phi[z^l_d] - \phi[x^l_n]\big\rVert_2 + \big\lVert\phi[z^r_d] - \phi[x^r_n]\big\rVert_2\Big],\label{eqn_str_pre_T_n}
\end{equation}	
where $\phi[x^l_n]$ denotes the feature map extracted from the `Conv4\_2' layer of the finetuned VGG-16 model for $x^l_n$, and so on. For the day$\rightarrow\allowbreak$night cycle, $\mathcal{L}_\text{str-pre}(T_d)$ is obtained from Eq.~(\ref{eqn_str_pre_T_n}) by swapping the domain labels from night to day. As shown in Fig.~\ref{figure_our_model2}a, the structure-preservation constraint is effective in handling fakeness for glow/glare regions, resulting more correct depths.

\subsubsection{Weighted-Smoothness Loss}
To handle the uninformative regions (low-light and saturated regions), we propose to add a weighted-smoothness constraint. The basic idea is that, in the uninformative regions with fake depths, we encourage the network to propagate depths from the neighbouring regions which have sufficient information and more accurate depths. Unlike the structure-preservation loss that is included in the training stage, we apply the weighted-smoothness loss in the testing stage (test-time optimisation). 

Given the input test pair $(x^l_n, x^r_n)$, we apply our weighted-smoothness constraint on the corresponding rendered image pair $(z^l_d, z^r_d)$:
\vspace{-0.05in}
\begin{align}
\thickmuskip=0mu
\mathcal{L}_\text{smooth}(T_n)\!=\!\mathbb{E}_{(x^l_n, x^r_n)}\Big[&\big\lVert w^l_n\big(|\partial_\text{x}(z^l_d)|\!+\!|\partial_\text{y}(z^l_d)|\big)\big\rVert_1 +\nonumber\\ 
&\big\lVert w^r_n\big(|\partial_\text{x}(z^r_d)|\!+\! |\partial_\text{y}(z^r_d)|\big)\big\rVert_1\Big],\label{eqn_smooth_T_n}
\end{align}
where the functions $\partial_\text{x}$ and $\partial_\text{y}$ compute the horizontal and vertical gradients respectively. $w^l_n$ is a weight map obtained from the input image $x^l_n$ using: $w^l_n\!=\!\exp\left(\frac{-([x^l_n]_{\max}-10)}{15}\right)+\exp\left(\frac{[x^l_n]_{\min}-250}{15}\right)$, where $[x^l_n]_{\max}$ and $[x^l_n]_{\min}$ are obtained from $x^l_n$ by taking the maximum and minimum pixel intensity among its three color channels. $w^r_n$ is similarly obtained from $x^r_n$. 
Thus in the weight maps, we assign higher values to the low-light and saturated regions, which enforces more smoothness in these regions. 
The results can be observed in Fig.~\ref{figure_our_model3}a, where we can see that with the weighted-smoothness constraint, the rendered day pair have less fake structures in the low-light regions and the estimated depths are more correct.

\subsection{Overall Losses}
As a summary of our method, in the training stage, our combined loss function is expressed as:
\vspace{-0.05in}
\begin{align}
\mathcal{L}_\text{All}(T_d, T_n, &D_d, D_n, S_n) = \mathcal{L}_\text{GAN}(T_d) + \mathcal{L}_\text{GAN}(T_n)\nonumber\\
&+\mathcal{L}_\text{GAN}(D_d) +\mathcal{L}_\text{GAN}(D_n)\nonumber\\
&+\lambda_\text{cyc}\big(\mathcal{L}_\text{cyc-con}(T_d)
+ \mathcal{L}_\text{cyc-con}(T_n)\big)\nonumber\\
&+\lambda_\text{str}\big(\mathcal{L}_\text{str-pre}(T_d)
+ \mathcal{L}_\text{str-pre}(T_n)\big)\nonumber\\
&+ \lambda_\text{ste}\big(\mathcal{L}_\text{ste-con}(T_d, S_n) 
+\mathcal{L}_\text{ste-con}(T_n)\big),\label{eqn_overall_loss}
\end{align}
where $\lambda_\text{cyc}$, $\lambda_\text{str}$, and $\lambda_\text{ste}$ are the scaling weights of the cycle-consistency, structure-preservation and stereo-consistency loss functions respectively. 
Minimizing Eq.~(\ref{eqn_overall_loss}) gives us the optimized networks $T_d^*$, $T_n^*$ and $S_n^*$. Then, in the testing stage, we finetune the network $T_n^*$ on the testing night pair $(x^l_n, x^r_n)$ using our weighted-smoothness loss, $\mathcal{L}_\text{smooth}(T_n)$, expressed in Eq.~(\ref{eqn_smooth_T_n}), which gives us $T_n^{**}$. 
As a result, we can get the final disparity result on the night image pair $(x^l_n, x^r_n)$ by applying  $S^*_n[T^*_d[T^{**}_n[(x^l_n, x^r_n)]]]$, which generates the predicted disparity map $\{{{y}^*_3}^l_n\}$.

\section{Experiments}
In all our experiments, we use images resized to 256$\times$512 resolution and hyper-parameters $\{w_1, w_2, w_3, \lambda_\text{cyc}, \lambda_\text{str}, \lambda_\text{ste}\}\!=\!\{0.5, 0.7, \allowbreak1, 10, 1, 0.05\}$. In the training stage, we use a batch size of 4 and set the number of training epochs to 40. We keep a constant learning rate of 0.0002 for all the networks for the first-half of the training epochs, and then linearly decay it to zero over the second-half. In the testing stage, during the finetuning of the network on the input test night pair using the weighted-smoothness constraint, we use 10 iterations\footnote{The number of iterations in the testing stage is decided by selecting the iteration that provides the best depth estimation performance on the validation night pairs.} and a constant learning rate of 0.00005. For depth evaluation, we use the bad-pixels percentage metric (from ~\cite{menze2015object}) governed by its error threshold parameter $\delta$; and, smaller numbers imply better depth estimation. 

\begin{table}[t!]
	\small
	\centering
	\renewcommand{\arraystretch}{1.2}
	\caption {Quantitative results on the SYNTHIA night test data.} \label{table_1_synthia_results}
	\begin{tabularx}{\columnwidth}{ c|Y|Y|Y }
		\toprule
		\multirow{2}{*}{}  & $\delta = 1$ & $\delta = 2$ & $\delta = 3$\\
		\midrule	 
		PSMNet(S$_\text{D}$) &36.16 &26.83 &21.48\\
		\hline
		CycleGAN + PSMNet(S$_\text{D}$) &30.28 &22.65 &18.25\\
		\hline
		Joint-SS~\cite{sharma2018into} &29.42 &22.16 &16.39\\
		\hline
		\cellcolor{gray!25}\textbf{Our Method} &\cellcolor{gray!25}\textbf{14.23} &\cellcolor{gray!25}\textbf{8.92} &\cellcolor{gray!25}\textbf{6.48} \\
		\hline
		\hline
		PSMNet(S$_\text{N}$) &5.24 &2.97 &2.13 \\
		\bottomrule
	\end{tabularx}	
\end{table}

\begin{figure}[t!]
	\vspace{-0.1in}
	\centering
	\captionsetup[subfigure]{labelformat=empty}
	\captionsetup[subfloat]{farskip=2pt}
	\subfloat[Input Night Image]{\includegraphics[width=0.235\textwidth]{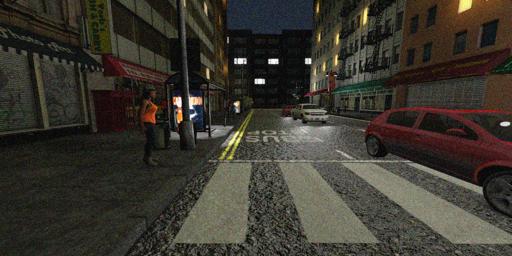}}\hfill
	\subfloat[ Disparity GT]{\includegraphics[width=0.235\textwidth]{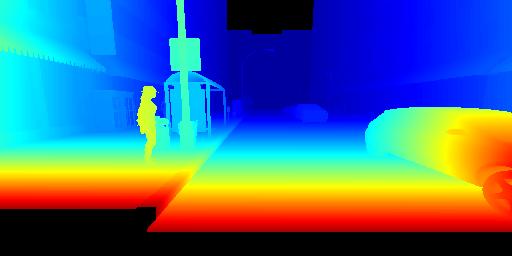}}\hfill\\
	\subfloat[ PSMNet(S$_\text{D}$)]{\includegraphics[width=0.235\textwidth]{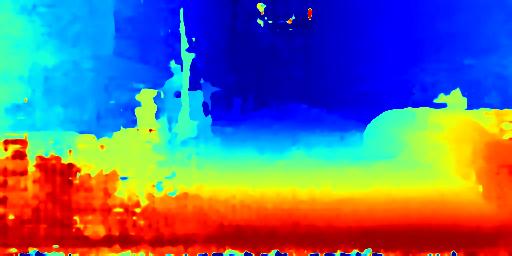}}\hfill
	\subfloat[\textbf{Our Result}]{\includegraphics[width=0.235\textwidth]{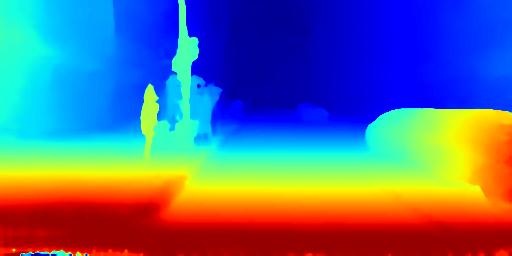}}\hfill\\
	\caption{Qualitative results on the SYNTHIA night test data. Our results are sharper and more accurate than the baseline methods.}\label{figure_synthia_results}
	\vspace{-0.1in}
\end{figure}

\begin{figure*}[t!]
	\vspace{-0.1in}
	\centering
	\captionsetup[subfigure]{labelformat=empty}
	\subfloat{\includegraphics[width=0.197\textwidth]{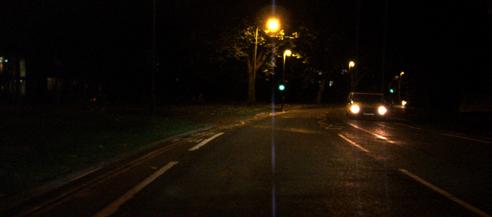}}\hfill   
	\subfloat{\includegraphics[width=0.197\textwidth]{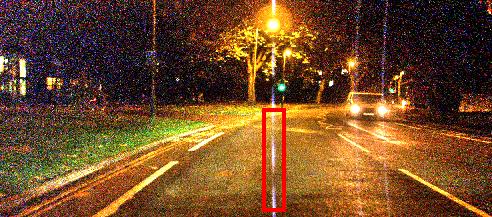}}\hfill
	\subfloat{\includegraphics[width=0.197\textwidth]{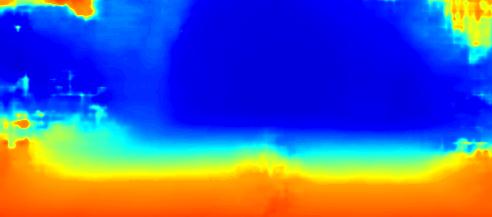}}\hfill
	\subfloat{\includegraphics[width=0.197\textwidth]{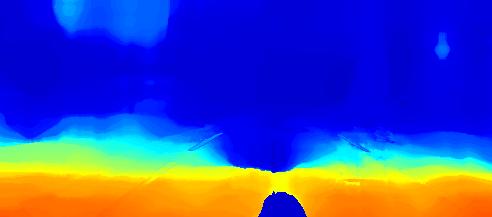}}\hfill
	\subfloat{\includegraphics[width=0.197\textwidth]{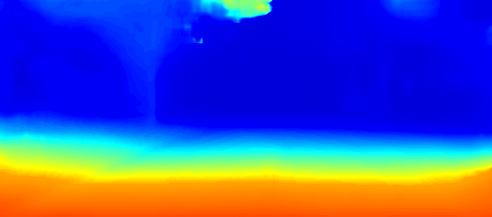}}\hfill\\
	\vspace{-0.125in}                                                              
	\subfloat{\includegraphics[width=0.197\textwidth]{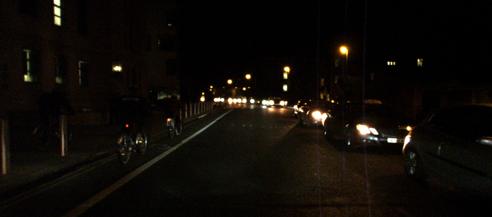}}\hfill
	\subfloat{\includegraphics[width=0.197\textwidth]{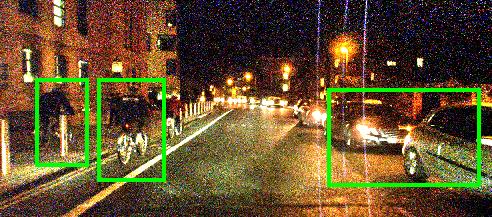}}\hfill
	\subfloat{\includegraphics[width=0.197\textwidth]{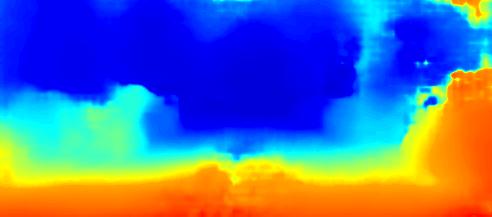}}\hfill
	\subfloat{\includegraphics[width=0.197\textwidth]{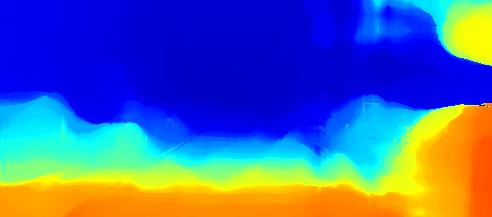}}\hfill
	\subfloat{\includegraphics[width=0.197\textwidth]{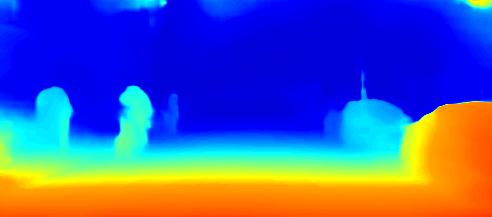}}\hfill\\
	\vspace{-0.125in}                                                               
	\subfloat[Input Night Image]{\includegraphics[width=0.197\textwidth]{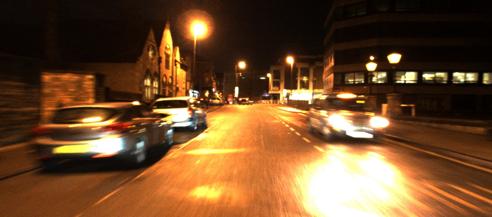}}\hfill
	\subfloat[Boosted Night Image]{\includegraphics[width=0.197\textwidth]{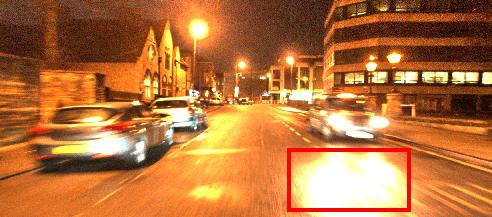}}\hfill
	\subfloat[ PSMNet(O$_\text{D}$)]{\includegraphics[width=0.197\textwidth]{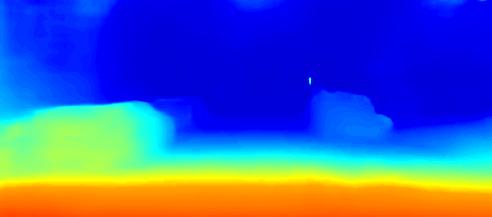}}\hfill
	\subfloat[ Joint-SS~\cite{sharma2018into}]{\includegraphics[width=0.197\textwidth]{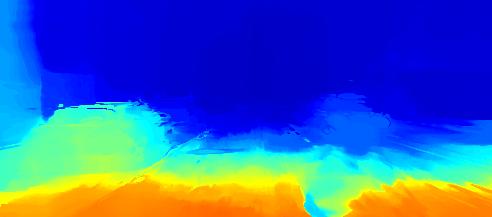}}\hfill
	\subfloat[\textbf{Our Result}]{\includegraphics[width=0.197\textwidth]{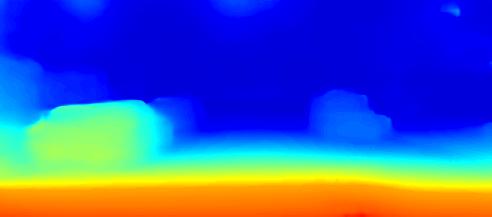}}\hfill\\
	\caption{Qualitative results on the Oxford night test data. Our results are robust to problems such as flares and high glow/glare (see the areas marked in red in the boosted night images). Additionally, they have sharper depth discontinuities (observe the cyclists, pedestrians, cars, etc. marked in green in the boosted night images). Note that, the boosted night images are shown for the sake of visualization.}\label{figure_oxford_results}
	\vspace{-0.1in}
\end{figure*}

\subsection{Results on Synthetic Night Data}
For our first experiment, we use the SYNTHIA~\cite{Ros_2016_CVPR} dataset, which provides day and night synthetic stereo data, and accurate dense disparity ground-truths. However, it does not realistically represent the complex attributes of real night images such as noise, flares, glow/glare. For this reason, we only use the methodology described in Sec.~\ref{subsec_learning_nighttime_depth} (i.e., without structure-preservation and weighted-smoothness constraints). We take 3800 stereo pairs (3500 for training and validation; 300 for testing) from the two data each, and add random Gaussian noise (st.d. $\in$ [0.04, 0.1]) in the night pairs to roughly simulate the low SNR conditions present in  real night images.

For comparisons, we select the following baselines: 1) `PSMNet(S$_\text{D}$)', PSMNet~\cite{chang2018pyramid} trained on the SYNTHIA day  data. 2) `CycleGAN + PSMNet(S$_\text{D}$)', CycleGAN~\cite{zhu2017unpaired} to translate the SYNTHIA night images into day images, which are then used with `PSMNet(S$_\text{D}$)'. 3) Joint-SS~\cite{sharma2018into}, a state-of-the-art method for nighttime stereo vision. And, 4) `PSMNet(S$_\text{N}$)', PSMNet~\cite{chang2018pyramid} trained on the SYNTHIA night data. Note that, we show the results of `PSMNet(S$_\text{N}$)' only to observe how our method performs with respect to the ``ideal'' case. Qualitative results are shown in Fig.~\ref{figure_synthia_results}. Quantitative results are presented in Table~\ref{table_1_synthia_results}. From the results, we can observe that our method generates disparity results which are sharper and more accurate than the baseline methods (see the supplement for additional results and comparisons with the baseline methods). 

\subsection{Results on Real Night Data} We test our model on real night data: the Oxford dataset \cite{maddern20171}, which provides a large amount of stereo data captured in daytime and nighttime conditions, and also provides their raw (LIDAR-based) sparse disparity maps. However, given the various limitations of LIDAR, these disparity maps are not suitable for training, until manually processed (as done in \cite{menze2015object}). Therefore, for pre-training $S_d$, we start with the KITTI \cite{menze2015object} day dataset that provides 200 day stereo pairs with their processed disparity ground-truth maps. We then finetune it on the Oxford day stereo data using the zoom-and-learn trick \cite{pang2018zoom} that provides finer and more accurate disparity maps for the Oxford day data. We take 9,000 day and 8,600 night pairs for training, and additional 1,000 night pairs for validation. We also ensure that the night pairs are diverse and consist of poorly-lit to well-lit conditions, flares, low-to-high glow/glare, etc. For testing, we use 250 additional night pairs, 125 from poorly-lit scenes (referred as Oxford (P-L)), and the other 125 from well-lit scenes (referred as Oxford (W-L))\footnote{Our evaluation on the Oxford dataset is based on the assumption that the raw depth maps are “mostly” correct, i.e. they are accurate for most of the points in the scene (like for static objects such as buildings, roads, etc.)}.

\begin{table}[t!]
	\small
	\centering
	\thickmuskip=3mu
	\renewcommand{\arraystretch}{1.2}
	\caption {Quantitative results on the Oxford night test data.} \label{table_2_oxford_results}
	\begin{tabularx}{\columnwidth}{ c|Y|Y|Y|Y }
		\toprule
		\multirow{2}{*}{} & \multicolumn{2}{c|}{Oxford (P-L)} & \multicolumn{2}{c}{Oxford (W-L)}\\
		&$\delta\!=\!3$ &$\delta\!=\!5$ & $\delta\!=\!3$ & $\delta\!=\!5$\\
		\midrule		 
		PSMNet(O$_\text{D}$) &16.58 &7.17 &8.24 &3.82\\
		\hline
		CycleGAN + PSMNet(O$_\text{D}$) &39.96 &20.33 &20.97 &8.59\\
		\hline
		PSMNet(S$_\text{N}$) &35.17 &22.19 &9.36 &5.60\\
		\hline
		PSMNet(O$_\text{D}$-LL) &14.97 &6.62 &8.75 &4.15\\
		\hline
		Joint-SS~\cite{sharma2018into} &12.75 &5.09 &7.92 &2.88\\
		\hline
		\cellcolor{gray!25}\textbf{Our Method} &\cellcolor{gray!25}\textbf{7.24} &\cellcolor{gray!25}\textbf{3.15} &\cellcolor{gray!25}\textbf{6.35} &\cellcolor{gray!25}\textbf{2.81}\\
		\bottomrule
	\end{tabularx}	
\end{table}

\begin{figure*}[t!]
	\vspace{-0.1in}
	\centering
	\captionsetup[subfigure]{labelformat=empty}
	\captionsetup[subfloat]{farskip=2pt}
	\subfloat{\includegraphics[width=0.1635\textwidth]{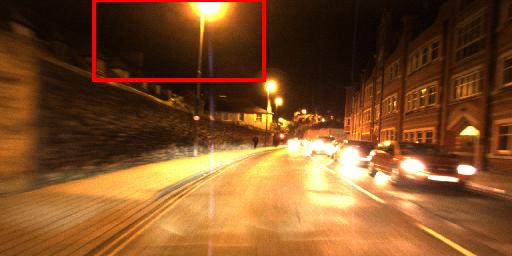}}\hfill
	\subfloat{\includegraphics[width=0.1635\textwidth]{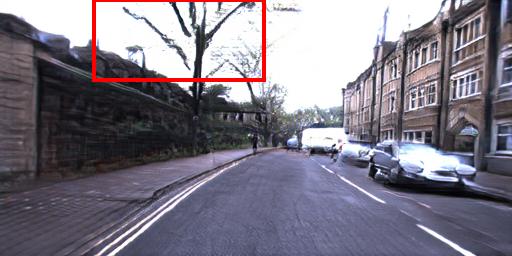}}\hfill
	\subfloat{\includegraphics[width=0.1635\textwidth]{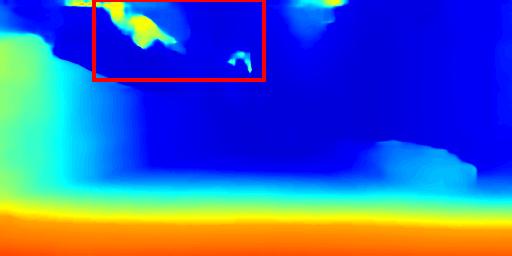}}\hfill
	\subfloat{\includegraphics[width=0.1635\textwidth]{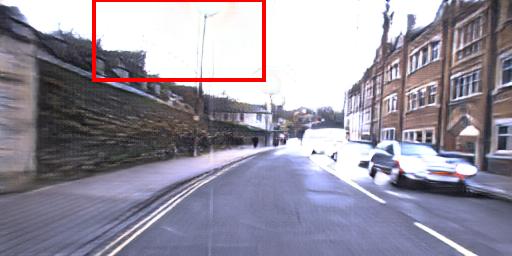}}\hfill
	\subfloat{\includegraphics[width=0.1635\textwidth]{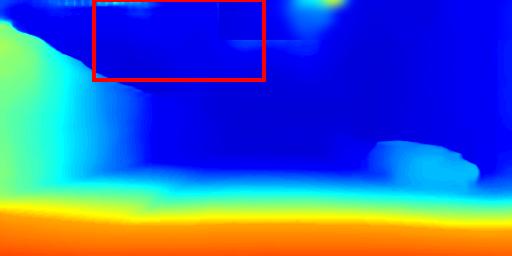}}\hspace{0.04in}
	\subfloat{\includegraphics[width=0.1635\textwidth]{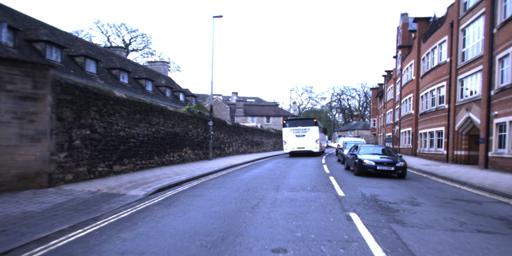}}\hfill\\
	\subfloat[Input Night Image]{\includegraphics[width=0.1635\textwidth]{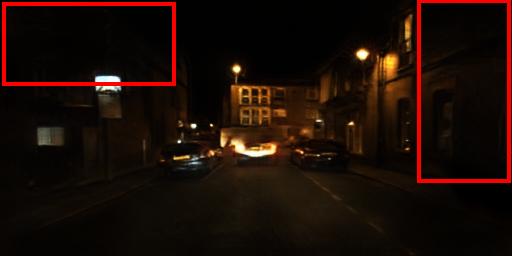}}\hfill
	\subfloat[Translation (w/o)]{\includegraphics[width=0.1635\textwidth]{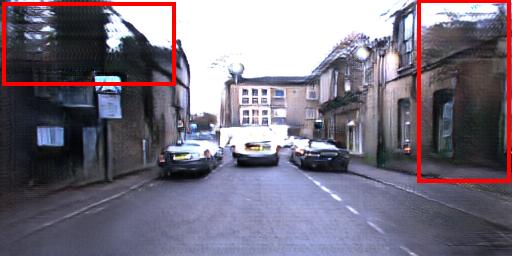}}\hfill
	\subfloat[Predicted Depth (w/o)]{\includegraphics[width=0.1635\textwidth]{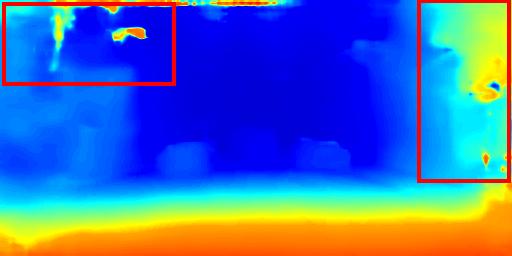}}\hfill
	\subfloat[\textbf{Translation} ]{\includegraphics[width=0.1635\textwidth]{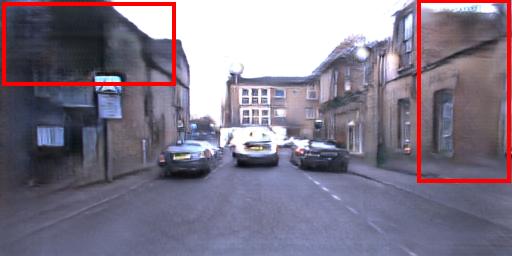}}\hfill
	\subfloat[\textbf{Predicted Depth}]{\includegraphics[width=0.1635\textwidth]{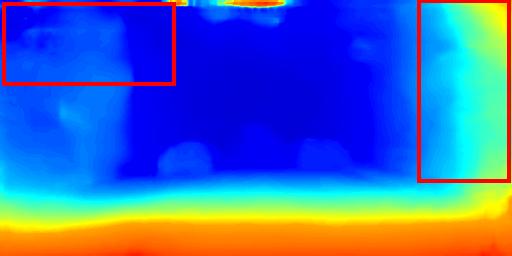}}\hspace{0.04in}
	\subfloat[Reference Day Image]{\includegraphics[width=0.1635\textwidth]{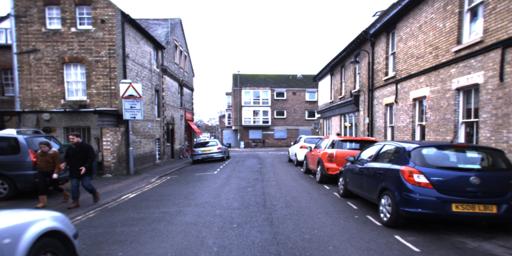}}\hfill\\       
	\caption{For the input night images, without the structure-preservation and weighted-smoothness constraints, the translated images have fake/improper structures (\nth{2} column) and the predicted depths are erroneous (\nth{3} column). However, when the constraints are used, the translated images (\nth{4} column) have more proper structures (for instance, we can observe that they have similar structures as in the reference day images (\nth{6} column) which roughly show the same scenes); and, the predicted depths (\nth{5} column) are also more correct. }\label{figure_fake_depths_sol}
	\vspace{-0.1in}
\end{figure*}

\begin{figure}[t]
	\vspace{-0.05in}
	\captionsetup[subfigure]{labelformat=empty}
	\captionsetup[subfloat]{farskip=2pt}
	\centering
	\subfloat{\includegraphics[width=0.156\textwidth]{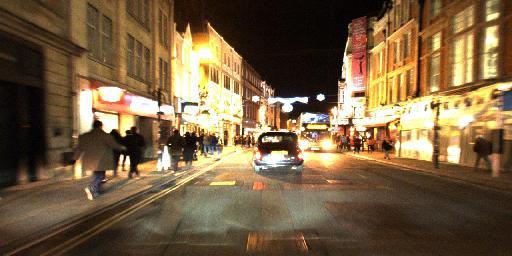}}\hfill
	\subfloat{\includegraphics[width=0.156\textwidth]{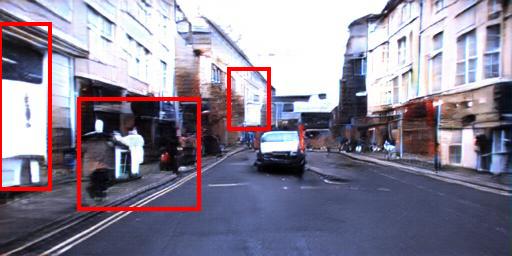}}\hfill
	\subfloat{\includegraphics[width=0.156\textwidth]{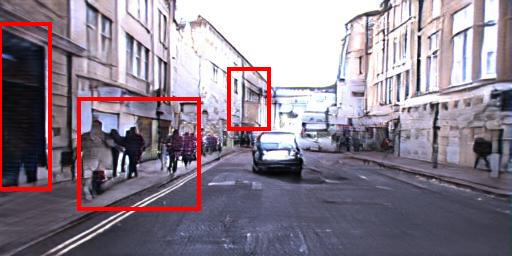}}\hfill\\
	\vspace{-0.05in}
	\setcounter{subfigure}{0}
	\subfloat[Input Night Pair]{\label{figure_motiv_stecons_1}\includegraphics[width=0.156\textwidth]{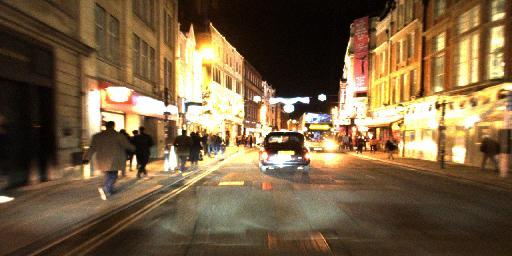}}\hfill
	\subfloat[ CycleGAN]{\label{figure_motiv_stecons_2}\includegraphics[width=0.156\textwidth]{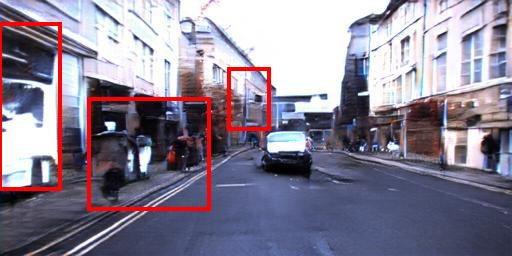}}\hfill
	\subfloat[ Ours]{\label{figure_motiv_stecons_3}\includegraphics[width=0.156\textwidth]{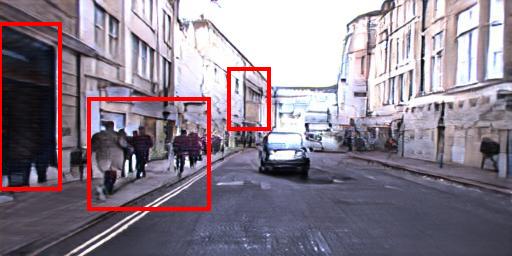}}\hfill\\
	\subfloat[ CycleGAN + PSMNet(O$_\text{D}$)]{\label{figure_motiv_stecons_4}\includegraphics[width=0.235\textwidth]{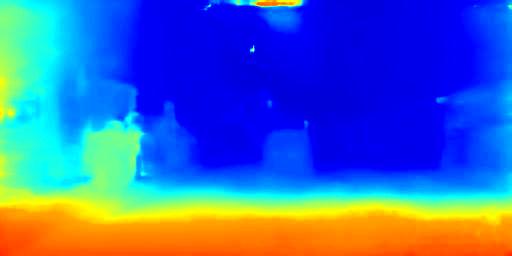}}\hfill
	\subfloat[ \textbf{Our Result}]{\label{figure_motiv_stecons_5}\includegraphics[width=0.235\textwidth]{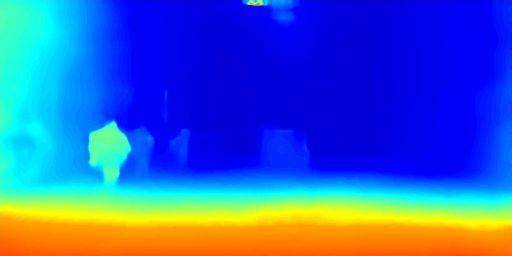}}\hfill
	\caption{Example showing the benefits of using our stereo-consistency constraint. We can observe that, compared to CycleGAN, our translated pair has lesser inconsistencies (compare the areas marked in red), and our depth result is more robust.}
	\label{figure_motiv_stecons}
	\vspace{-0.125in}
\end{figure}

For comparison, we have the following baselines: 1) `PSMNet(O$_\text{D}$)' , PSMNet~\cite{chang2018pyramid} finetuned on the Oxford day data (as discussed above). 2) `CycleGAN + PSMNet(O$_\text{D}$)', CycleGAN~\cite{zhu2017unpaired} to translate the Oxford night images into day images, which are then used with `PSMNet(O$_\text{D}$)'. 3) `PSMNet(S$_\text{N}$)'. 
4) `PSMNet(O$_\text{D}$-LL)' where we finetune PSMNet on the synthetic low-light Oxford  data which is created by randomly lowering the contrast of the Oxford day images to $2\!-\!4$\% of their original contrast, and then adding random Gaussian noise to them. 
And, 5) Joint-SS~\cite{sharma2018into}, a state-of-the-art method for night stereo vision. Quantitative results are shown in Table ~\ref{table_2_oxford_results}. Qualitative comparisons are shown in Fig.~\ref{figure_oxford_results}.

We can observe that compared to ours, `PSMNet(S$_\text{N}$)' performs poorly on real night images. This is due to the domain gap problem that arises on training on synthetic night data and testing using real night data. A similar domain gap problem problem exists for `PSMNet(O$_\text{D}$)' and `PSMNet(O$_\text{D}$-LL)'. Since `PSMNet(O$_\text{D}$)' is trained on day images, it cannot handle problems such as dense-noise, flares, etc. present in real night images, and produces blurry and incorrect results. `PSMNet(O$_\text{D}$-LL)' though performing better than `PSMNet(O$_\text{D}$)' on poorly-lit images, is also inferior to our method, since it is trained on synthetic low-light data which is again not a real representation of the real night data. We obtain a better performance than the individual approach, `CycleGAN + PSMNet(O$_\text{D}$)', which highlights the benefits of our joint network using our stereo-consistency constraint (see Fig.~\ref{figure_motiv_stecons} for an example). Our results are stable under a variety of nighttime conditions such as poorly-lit (first two rows, Fig.~\ref{figure_oxford_results} ) to well-lit scenes (last row, Fig.~\ref{figure_oxford_results}). Our results are more accurate and have sharper boundaries than the baseline methods (see the marked areas in green in the boosted night images, Fig.~\ref{figure_oxford_results}), They are robust to flares and high glow/glare (see the marked areas in red in the boosted night images, Fig.~\ref{figure_oxford_results}). 
Fig.~\ref{figure_fake_depths_sol} shows some examples demonstrating the effectiveness of our method in mitigating the fake depth problem for glow/glare and uninformative regions. See the supplement for more results.

\section{Ablation Studies}
\begin{figure}[t!]
	\vspace{-0.1in}
	\centering
	\captionsetup[subfigure]{labelformat=empty}
	\captionsetup[subfloat]{farskip=2pt}
	\subfloat{\includegraphics[width=0.235\textwidth]{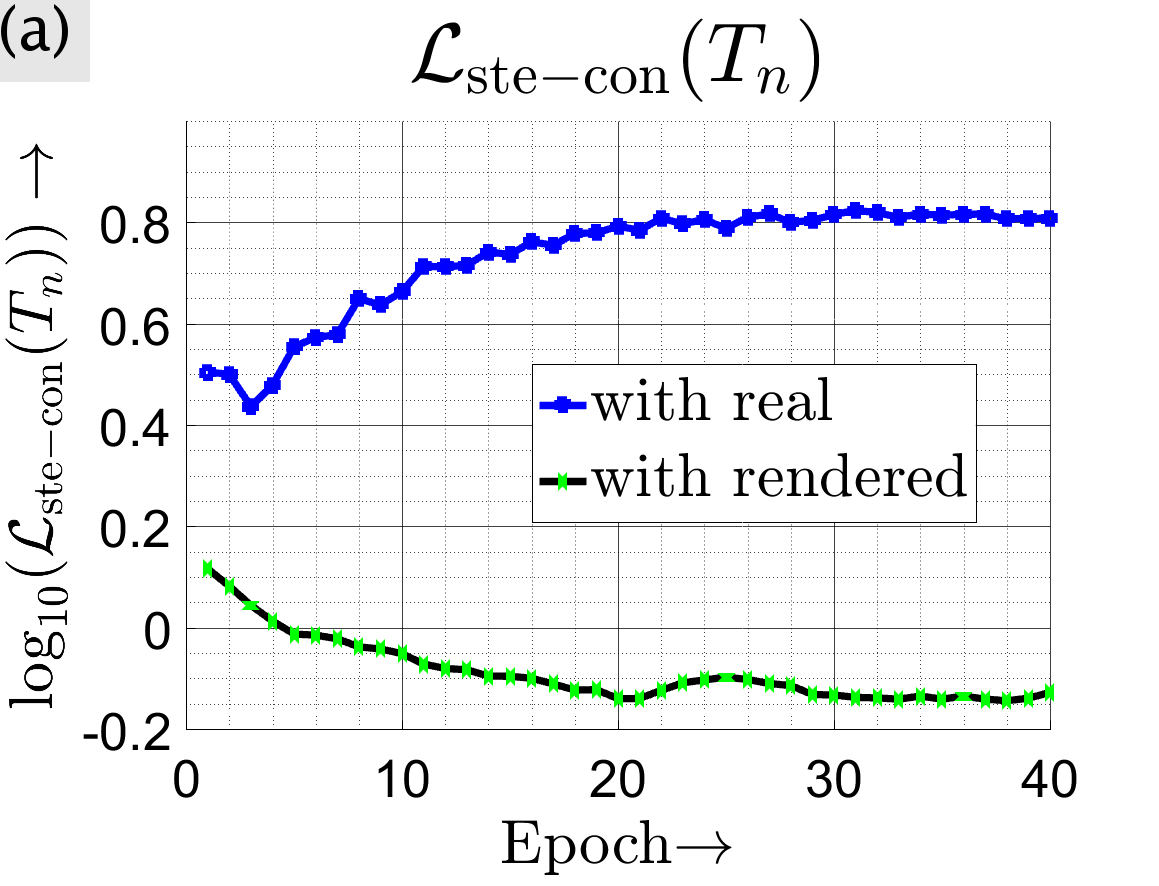}}\hfill
	\subfloat{\includegraphics[width=0.235\textwidth]{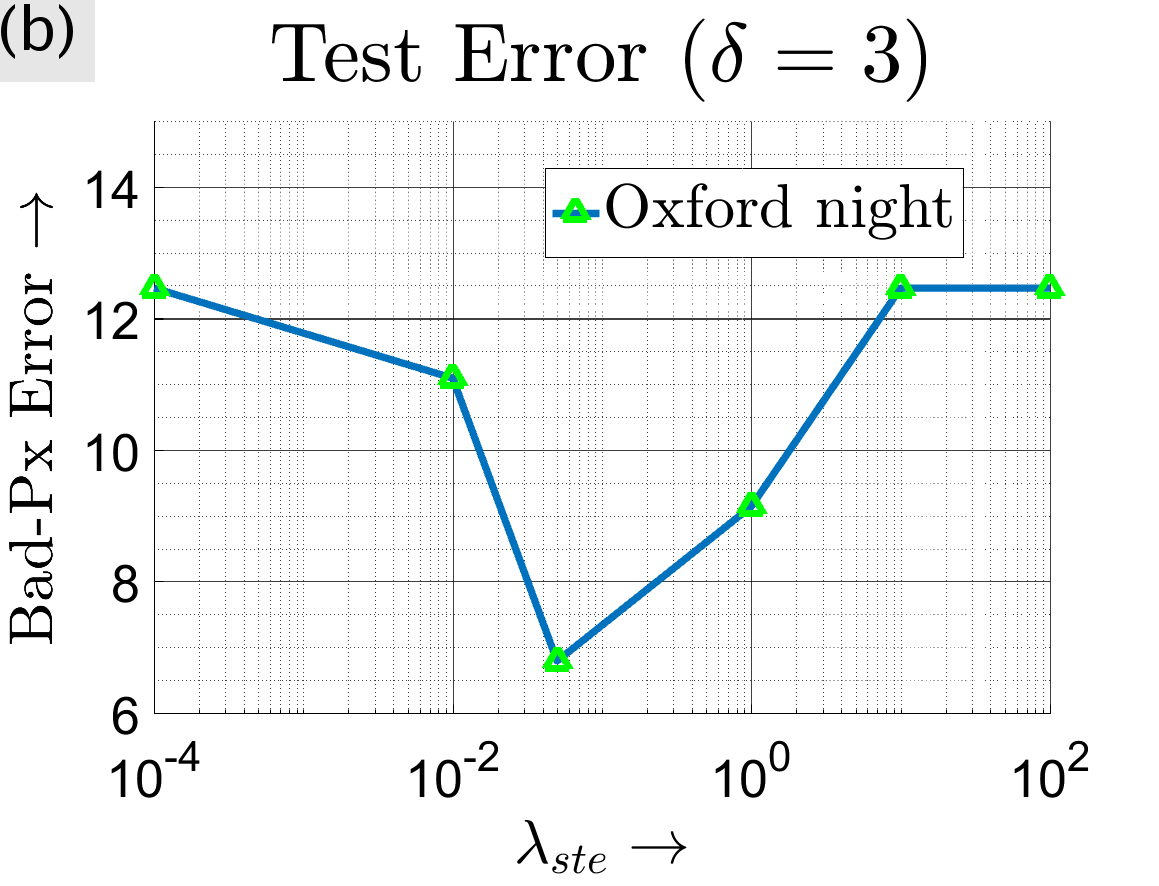}}\hfill    
	\caption{(a) Training stabilizes and the stereo-consistency loss $\mathcal{L}_\text{ste-con}(T_n)$ converges when rendered night pairs are used instead of real night pairs (Eq.~(\ref{eqn_ste_con_T_n})). (b) Test error w.r.t. $\lambda_\text{ste}$, parameter controlling the effect of the stereo-consistency constraint.}\label{figure_ablation}
	\vspace{-0.1in}
\end{figure}

\paragraph{Stereo-Consistency Constraint} 
Fig.~\ref{figure_ablation}a shows training stabilization when rendered night pairs are used instead of real night pairs for the stereo-consistency loss $\mathcal{L}_\text{ste-con}(T_n)$ in Eq.~(\ref{eqn_ste_con_T_n}). Fig.~\ref{figure_ablation}b shows the variation of the least test error on the Oxford night test data with $\lambda_\text{ste}$, the parameter controlling the strength of the stereo-consistency constraint. The results shown in Figs.~\ref{figure_motiv_stecons} and \ref{figure_ablation}b confirm the efficacy of using this constraint in our method. Also, $\lambda_\text{ste}$ should be chosen moderately and it should neither be too low (constraint is too weak) nor too high (constraint is too strong).

\paragraph{Structure-Preservation and Weighted-Smoothness Constraints}
For the structure-preservation constraint, using a pre-trained VGG-16 model instead of our finetuned VGG-16 model increases our method's test error ($\delta$=3) to 7.07 from 6.35 on the Oxford (W-L) test set. Also, without the constraint, the error  increases to 7.71 (i.e., our performance drops by $\sim$21\%). As for the weighted-smoothness constraint, without using it, our method's test error increases to 9.35 from 
7.24 on the Oxford (P-L) test set (i.e., our performance drops by $\sim$29\%). The results show that both the constraints are important for our method's better performance, as they help in mitigating the fake depth problem (Sec.~\ref{subsec_handling_fake_depths}).

\section{Conclusion}
In this paper, to address the problem of nighttime stereo depth estimation, we have proposed a joint translation-stereo network that does not require disparity ground-truths of the night training images. We have included a stereo-consistency constraint to ensure that the translated image pairs are stereo-consistent, which is imperative for stereo matching. In addition, to handle the fake depths problem that occur due to the unsupervised/unpaired learning of the translation process, we have added structure-preservation and weighted-smoothness constraints to mitigate the fake depths problem for regions with glow/glare light effects and uninformative regions. 

\newpage

{\small
	\bibliographystyle{ieee}
	\bibliography{egbib}

\begin{thebibliography}{10}\itemsep=-1pt

\bibitem{hoffman2017cycada}
Cycada: Cycle consistent adversarial domain adaptation.
\newblock In {\em International Conference on Machine Learning (ICML)}, 2018.

\bibitem{chang2018pyramid}
J.-R. Chang and Y.-S. Chen.
\newblock Pyramid stereo matching network.
\newblock In {\em Proceedings of the IEEE Conference on Computer Vision and
  Pattern Recognition}, pages 5410--5418, 2018.

\bibitem{guo2017lime}
X.~Guo, Y.~Li, and H.~Ling.
\newblock Lime: Low-light image enhancement via illumination map estimation.
\newblock {\em IEEE Transactions on Image Processing}, 26(2):982--993, 2017.

\bibitem{heo2007simultaneous}
Y.~S. Heo, K.~M. Lee, and S.~U. Lee.
\newblock Simultaneous depth reconstruction and restoration of noisy stereo
  images using non-local pixel distribution.
\newblock In {\em Computer Vision and Pattern Recognition, 2007. CVPR'07. IEEE
  Conference on}, pages 1--8. IEEE, 2007.

\bibitem{hirschmuller2005accurate}
H.~Hirschmuller.
\newblock Accurate and efficient stereo processing by semi-global matching and
  mutual information.
\newblock In {\em Computer Vision and Pattern Recognition, 2005. CVPR 2005.
  IEEE Computer Society Conference on}, volume~2, pages 807--814. IEEE, 2005.

\bibitem{jiao2017joint}
J.~Jiao, Q.~Yang, S.~He, S.~Gu, L.~Zhang, and R.~W. Lau.
\newblock Joint image denoising and disparity estimation via stereo structure
  pca and noise-tolerant cost.
\newblock {\em International Journal of Computer Vision}, 124(2):204--222,
  2017.

\bibitem{johnson2016perceptual}
J.~Johnson, A.~Alahi, and L.~Fei-Fei.
\newblock Perceptual losses for real-time style transfer and super-resolution.
\newblock In {\em European Conference on Computer Vision}, pages 694--711.
  Springer, 2016.

\bibitem{jolicoeur2018relativistic}
A.~Jolicoeur-Martineau.
\newblock The relativistic discriminator: a key element missing from standard
  gan.
\newblock {\em arXiv preprint arXiv:1807.00734}, 2018.

\bibitem{kendall2017end}
A.~Kendall, H.~Martirosyan, S.~Dasgupta, P.~Henry, R.~Kennedy, A.~Bachrach, and
  A.~Bry.
\newblock End-to-end learning of geometry and context for deep stereo
  regression.
\newblock {\em CoRR, vol. abs/1703.04309}, 2017.

\bibitem{laga2020survey}
H.~Laga, L.~V. Jospin, F.~Boussaid, and M.~Bennamoun.
\newblock A survey on deep learning techniques for stereo-based depth
  estimation.
\newblock {\em arXiv preprint arXiv:2006.02535}, 2020.

\bibitem{liu2017unsupervised}
M.-Y. Liu, T.~Breuel, and J.~Kautz.
\newblock Unsupervised image-to-image translation networks.
\newblock In {\em Advances in Neural Information Processing Systems}, pages
  700--708, 2017.

\bibitem{ExDark}
Y.~P. Loh and C.~S. Chan.
\newblock Getting to know low-light images with the exclusively dark dataset.
\newblock {\em Computer Vision and Image Understanding}, 178:30--42, 2019.

\bibitem{maddern20171}
W.~Maddern, G.~Pascoe, C.~Linegar, and P.~Newman.
\newblock 1 year, 1000 km: The oxford robotcar dataset.
\newblock {\em The International Journal of Robotics Research}, 36(1):3--15,
  2017.

\bibitem{mao2017least}
X.~Mao, Q.~Li, H.~Xie, R.~Y. Lau, Z.~Wang, and S.~P. Smolley.
\newblock Least squares generative adversarial networks.
\newblock In {\em Computer Vision (ICCV), 2017 IEEE International Conference
  on}, pages 2813--2821. IEEE, 2017.

\bibitem{menze2015object}
M.~Menze and A.~Geiger.
\newblock Object scene flow for autonomous vehicles.
\newblock In {\em Proceedings of the IEEE Conference on Computer Vision and
  Pattern Recognition}, pages 3061--3070, 2015.

\bibitem{pang2018zoom}
J.~Pang, W.~Sun, C.~Yang, J.~Ren, R.~Xiao, J.~Zeng, and L.~Lin.
\newblock Zoom and learn: Generalizing deep stereo matching to novel domains.
\newblock In {\em The IEEE Conference on Computer Vision and Pattern
  Recognition (CVPR)}, June 2018.

\bibitem{Ros_2016_CVPR}
G.~Ros, L.~Sellart, J.~Materzynska, D.~Vazquez, and A.~M. Lopez.
\newblock The synthia dataset: A large collection of synthetic images for
  semantic segmentation of urban scenes.
\newblock In {\em The IEEE Conference on Computer Vision and Pattern
  Recognition (CVPR)}, June 2016.

\bibitem{ImageNet}
O.~Russakovsky, J.~Deng, H.~Su, J.~Krause, S.~Satheesh, S.~Ma, Z.~Huang,
  A.~Karpathy, A.~Khosla, M.~Bernstein, A.~C. Berg, and L.~Fei-Fei.
\newblock {ImageNet Large Scale Visual Recognition Challenge}.
\newblock {\em International Journal of Computer Vision (IJCV)},
  115(3):211--252, 2015.

\bibitem{sharma2018into}
A.~Sharma and L.-F. Cheong.
\newblock Into the twilight zone: Depth estimation using joint structure-stereo
  optimization.
\newblock In {\em Proceedings of the European Conference on Computer Vision
  (ECCV)}, pages 103--118, 2018.

\bibitem{zhang2015meshstereo}
C.~Zhang, Z.~Li, Y.~Cheng, R.~Cai, H.~Chao, and Y.~Rui.
\newblock Meshstereo: A global stereo model with mesh alignment regularization
  for view interpolation.
\newblock In {\em Proceedings of the IEEE International Conference on Computer
  Vision}, pages 2057--2065, 2015.

\bibitem{flowdark}
Y.~Zheng, M.~Zhang, and F.~Lu.
\newblock Optical flow in the dark.
\newblock In {\em The IEEE/CVF Conference on Computer Vision and Pattern
  Recognition (CVPR)}, June 2020.

\bibitem{zhu2017unpaired}
J.-Y. Zhu, T.~Park, P.~Isola, and A.~A. Efros.
\newblock Unpaired image-to-image translation using cycle-consistent
  adversarial networks.
\newblock In {\em Computer Vision (ICCV), 2017 IEEE International Conference
  on}, 2017.

\end{thebibliography}
}

\end{document}